%% file: __main.tex
\documentclass[10.5pt]{article}
\input{packages}

\newcommand{\circled}[1]{\tikz[baseline=(char.base)]{
  \node[shape=circle,draw,inner sep=1pt] (char) {#1};
}}

\title{
LLM Inference in a Flash!
}

\author{
Sebastian Zhao\blfootnote{Equal contribution}$^{*1}$ \enspace\enspace Minseo Kim$^{*1}$ \enspace\enspace Coleman Hooper$^{*1}$ \enspace\enspace Luca Manolache$^{1}$\\
Michael W. Mahoney$^{1,2,3}$\enspace\enspace Yakun Sophia Shao$^{1}$\enspace\enspace Kurt Keutzer$^{1}$\enspace\enspace Amir Gholami$^{1,2}$\vspace{3mm}
\\
{$^{1}$~UC Berkeley\qquad $^{2}$~ICSI\qquad $^{3}$~LBNL\vspace{3mm}}\\
}
\date{}
\makeatother

\begin{document}

\maketitle

\input{_0_abstract}
\input{_1_intro}
\input{_2_related}
\input{_3_algorithm}
\input{_4_system_modeling}

\input{_5_results}
\input{_6_conclusion}

\section*{Acknowledgements}
We acknowledge the gracious support from the FuriosaAI, Intel, Apple, NVIDIA, Macronix, and Mozilla teams.
Furthermore, we appreciate the support from Google Cloud, the Google TRC team, Prof. David Patterson, along with support from the Google Gemini team and Divy Thakkar.
Prof. Keutzer's lab is also sponsored by funding through BDD and BAIR.
We also acknowledge support by the Director, Office of Science, Office of Advanced Scientific Computing Research, of the U.S. Department of Energy under Contract No. DE-AC02-05CH11231.
Michael W. Mahoney would also like to acknowledge DARPA, DOE, NSF, ONR, and the DOE SciGPT grant.
Our conclusions do not necessarily reflect the position or the policy of our sponsors, and no official endorsement should be~inferred.


\bibliographystyle{plain}
\bibliography{references}

\appendix
\counterwithin{figure}{section}
\counterwithin{table}{section}
\input{_7_appendix}

\end{document}

%% file: packages.tex
\usepackage{wrapfig}

\usepackage{mathpazo}
\usepackage[utf8]{inputenc}
\usepackage{verbatim}
\usepackage{calc}
\usepackage{mathrsfs}
\usepackage{mathtools}
\usepackage{url}
\usepackage{algorithm}
\usepackage{algpseudocode}
\usepackage{tikz}
\usepackage{amsmath}
\usepackage{amssymb}
\usepackage{graphicx}
\usepackage{hyperref}
\usepackage{xcolor}
\usepackage{comment}
\usepackage{soul}
\usepackage{booktabs}

\definecolor{forestgreen}{rgb}{0.13, 0.55, 0.13}

\usepackage{tabularx}

\makeatletter

\newcommand{\lyxmathsym}[1]{\ifmmode\begingroup\def\b@ld{bold}
  \text{\ifx\math@version\b@ld\bfseries\fi#1}\endgroup\else#1\fi}

\usepackage{graphics}\usepackage{epsfig}

\newcommand\hc{ \rowcolor{teal!10}}
\newcommand\hd{ \rowcolor{teal!18}}

\usepackage{soul}

\usepackage{graphicx}
\usepackage{booktabs} 

\usepackage{here}

\usepackage{amsmath,amssymb,amsfonts,amsbsy,amsfonts,latexsym}
\usepackage{multirow}
\usepackage{makecell}
\usepackage[labelfont=bf,belowskip=0pt,aboveskip=5pt,tableposition=top]{caption}
\usepackage{xcolor}
\usepackage{colortbl}

\definecolor{colorA}{RGB}{189,201,225}
\definecolor{colorB}{RGB}{103,169,207}
\definecolor{colorC}{RGB}{ 28,144,153}
\definecolor{colorD}{RGB}{  1,108, 89}

\newcolumntype{R}{>{\columncolor{gray!40}}r}
\newcolumntype{L}{>{\columncolor{gray!40}}l}
\newcolumntype{C}{>{\columncolor{gray!40}}c}

\newcommand\blfootnote[1]{%
  \begingroup
  \renewcommand\thefootnote{}\footnote{#1}%
  \addtocounter{footnote}{-1}%
  \endgroup
}

\usepackage{tabularx,colortbl,xcolor}
\usepackage{multirow}
\usepackage[normalem]{ulem}
\useunder{\uline}{\ul}{}

\usepackage{enumitem}

\usepackage{xparse}

\DeclareGraphicsExtensions{.pdf,.png}

\usepackage{longtable}
\usepackage{pgfplots}
\usepackage{outlines}

\usepackage[many]{tcolorbox}
\usepackage{caption}
\usepackage{subcaption}
\usepackage{graphbox} 

\tcbset{
    sharp corners,
    colback = white,
    before skip = 0.2cm,    
    after skip = 0.5cm      
}                           

\definecolor{main}{HTML}{4472C4}    
\definecolor{sub}{HTML}{EBF4FF}     

\newtcolorbox{boxA}{
    enhanced, breakable,
    boxrule = 0pt,
    colback = sub,
    borderline west = {2pt}{0pt}{main}, 
    borderline east = {2pt}{0pt}{main}, 
}

\usepackage{times}
\usepackage{textcomp}

%% file: _0_abstract.tex
\begin{abstract}
Large Language Models (LLMs) have shown impressive capabilities across a range of natural language processing tasks, and LLM inference has emerged as a critical workload for enabling downstream applications.
The demands of serving LLM inference are becoming increasingly challenging as requests shift toward longer sequences and heavier inference, driven by retrieval-augmented generation, inference-time compute scaling, and long-context applications.
Additionally, these challenges are compounded by hardware trends, as memory capacity and communication bandwidth are not scaling as fast as increases in workload complexity.
Compute-in-Flash is a promising solution to address memory bandwidth limitations by moving computation close to memory, and to exploit the large capacity of SSD technologies. However, it is challenging to deploy LLMs on these systems as they lack support for high-precision floating point operations and have limited write endurance.
In our work, we aim to address these challenges by designing inference algorithms to enable LLM inference on Flash compute-in-memory devices.
We present an end-to-end integer-only quantization approach to eliminate expensive floating-point computations.
To address the limited write endurance, we design a dictionary-based KV cache compression strategy based on sparse dictionary coding that represents each KV vector as a linear combination of static dictionary vectors.
These algorithmic improvements enable us to exploit the benefits of Compute-in-Flash for both model weights and KV cache, and to minimize expensive data transfer operations.
Across Llama-3.1-8B and Qwen-2.5-7B, our combined method exhibits limited accuracy degradation while reducing dynamic KV cache traffic by 15$\times$.
We then model the system-level benefits of our method, showing how it enables \textbf{3.1}$\times$/\textbf{2.7}$\times$ latency/energy savings for short context lengths and \textbf{4.4}$\times$/\textbf{6.8}$\times$ latency/energy savings for long context lengths relative to inference with a standard DRAM-enabled system.

\end{abstract}

%% file: _1_intro.tex
\section{Introduction}

Large Language Models (LLMs) have enabled groundbreaking progress in natural language processing capabilities.
With these advances, LLM inference has become a critical workload to enable a range of downstream applications like chatbots and agents~\cite{yao2022react, schick2023toolformer, durante2024agent}.
However, LLM inference is incredibly resource-intensive, with memory constraints emerging as a key bottleneck for inference due to substantial storage requirements for model weights as well as for cached activations for longer contexts~\cite{gholami2024ai}.
Autoregressive decoding is typically memory-bandwidth-bound because each token requires reading the growing KV cache, limiting parallelism~\cite{kim2024squeezellm}.
This pressure increases with longer contexts and serving patterns such as retrieval-augmented generation~\cite{lewis2020retrieval} and inference-time compute scaling~\cite{snell2024scaling}.

These workload trends are compounded by hardware scaling limits: memory capacity and bandwidth scale more slowly than compute, a phenomenon often referred to as the memory wall~\cite{gholami2024ai}.
This gap is further amplified by practical limits of SRAM/DRAM-based memory hierarchies.
SRAM-only designs are not able to provide sufficient capacity for LLM inference workloads~\cite{ma2026challenges}.
Additionally, energy consumption for memory transfers is not scaling down as we shift to newer technology nodes~\cite{jouppi2021ten}.
HBM and DRAM technologies are also not scaling as we go to newer compute nodes, and are becoming increasingly expensive due to limited supply~\cite{ma2026challenges}.
While Flash memory provides much greater capacities, these systems are typically unsuitable for LLM inference due to high energy consumption for off-chip memory transfers, as well as limited write endurance (which is unsuitable for KV cache activations)~\cite{yu2024cambricon, alizadeh2024llm}.

\begin{figure*}[t]
  \centering
  \includegraphics[width=0.99\textwidth]{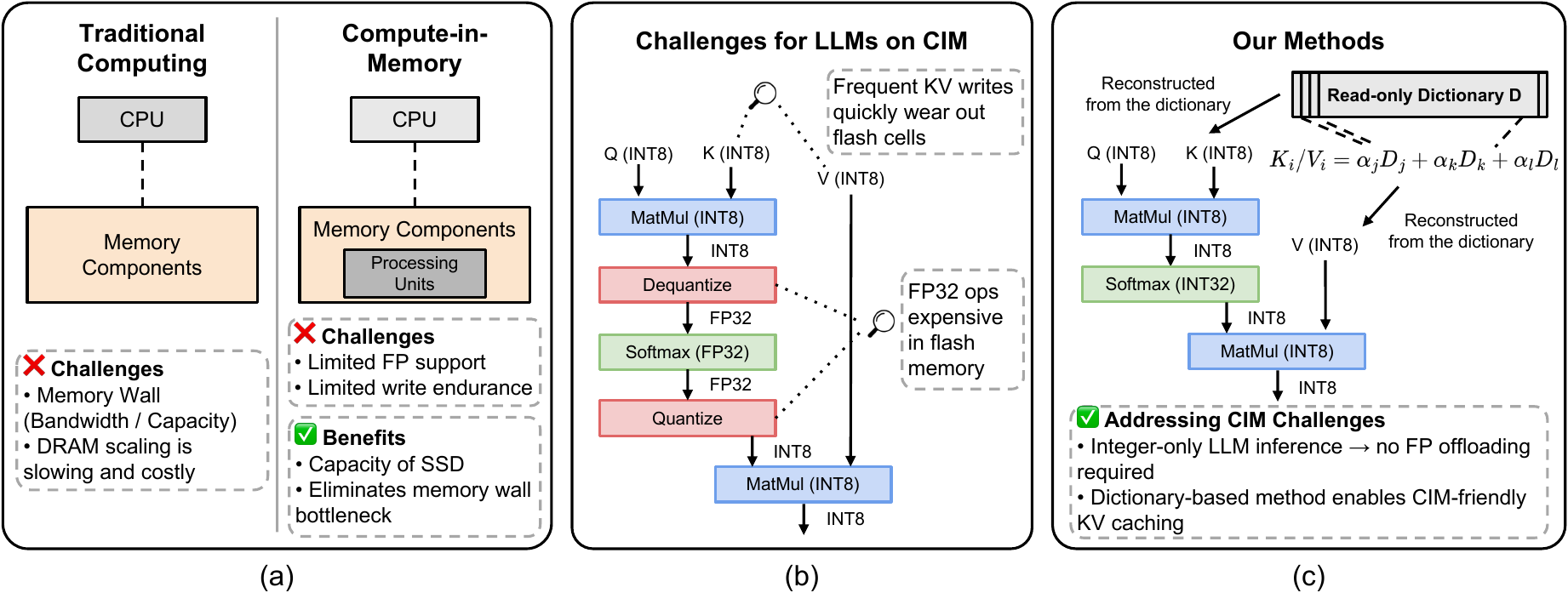}
  \caption{
  Motivation and overview of LLM inference on flash compute-in-memory (CIM).
  (a) Comparison between traditional computing and CIM, illustrating the memory-wall bottleneck.
  (b) Key challenges for LLM decoding on flash CIM: frequent KV-cache writes (endurance) and costly FP32/nonlinear operations.
  (c) Proposed methods: end-to-end integer-only inference and dictionary-based KV caching.
  }
  \vspace*{-3mm}
  \label{fig:teaser}
\end{figure*}

Compute-in-memory has emerged as a promising solution to address memory bandwidth limitations by moving computation close to memory (Figure~\ref{fig:teaser}(a))~\cite{hu2022ice, mutlu2023memory, lu2023rram}.
Compute-in-Flash can potentially allow computations to be performed directly in the memory array, thereby reducing energy and latency from data transfers~\cite{wolters2024memory}.
However, there are multiple challenges with deploying LLMs on Compute-in-Flash systems (Figure~\ref{fig:teaser}(b)).
These systems typically do not support high-precision floating-point operations (e.g. for nonlinearities in LLMs), have low off-chip bandwidth, and limited write endurance (which makes them unsuitable for dynamic activations)~\cite{ma2026challenges}.
There is a need for algorithmic improvements to enable LLM inference to be deployed on compute-in-memory technologies.

In this work, we enable LLM inference on Compute-in-Flash systems.
We first design a quantization strategy that supports LLM inference on devices without efficient floating-point arithmetic.
This strategy uses integer arithmetic end-to-end, including for nonlinear operations.
To address the large, dynamic KV cache, which is difficult to place on flash due to limited write endurance, we develop a KV-cache compression strategy that leverages a large read-only dictionary.
This approach enables KV reconstruction while substantially reducing dynamic state and write traffic.
Together, our techniques map both weights and KV-cache operations to Compute-in-Flash and minimize off-chip data transfers and dynamic memory operations.

Specifically, we make the following contributions:
\begin{itemize}
    \item 
    We design an end-to-end integer-only quantization strategy that eliminates floating-point computation throughout LLM inference, including attention and nonlinear operations.
    This enables efficient execution on Compute-in-Flash devices and reduces latency and energy by avoiding FP offloading.
    \item
    We develop a dictionary-based KV-cache compression method based on sparse dictionary coding, representing each KV vector with sparse codes over a static, flash-resident dictionary.
    This design minimizes write traffic while exploiting flash capacity and internal read bandwidth, with only lightweight metadata handled by the host.
    \item
    We present a system-level analytical model that quantifies the latency and energy benefits of our design under realistic Compute-in-Flash constraints.
    Our model shows how our approach enables \textbf{3.1}$\times$/\textbf{2.7}$\times$ latency/energy savings for short context lengths and \textbf{4.4}$\times$/\textbf{6.8}$\times$ latency/energy savings for long context lengths (for our S2 configuration; see Section~\ref{subsec:results-system}), enabling efficient LLM inference on resource-constrained systems.
\end{itemize}

%% file: _2_related.tex
\section{Related Work}

\subsection{Compute-in-Memory for LLM Inference}

There has been existing work on compute-in-memory processing for LLM inference~\cite{wolters2024memory,zhu2025leveraging}. A recent survey provides an overview of compute-in-memory technologies for LLM inference~\cite{ma2026challenges}. In particular, several existing works have aimed to exploit Compute-in-Flash technologies to minimize data transfers~\cite{kim2025llm,ji2026flash,lee2025aif}.
LLM-on-the-Palm~\cite{kim2025llm} stores weights in Compute-in-Flash for efficient processing on resource-constrained mobile devices with limited DRAM bandwidth.
AiF~\cite{lee2025aif} tailors the Flash chip for LLM inference by enabling highly concurrent read accesses.
Other work aims to support KV caching in compute-in-memory for long-context inference~\cite{gu2025pim}.

Key challenges for running LLM inference on Flash technologies include limited support for high precision floating point operations, limited write endurance, and low off-chip bandwidth~\cite{ma2026challenges}, as well as substantial energy requirements for transferring data to/from the Flash chip \cite{kyung2025ssd}. There has been existing work on co-designing LLM inference to adapt to the limitations of Flash computing technologies.
LLM-in-a-Flash~\cite{alizadeh2024llm} stores model weights in SSDs to leverage their capacity and adapts the inference algorithm to minimize data transfers under limited SSD bandwidth.
Cambricon-LLM~\cite{yu2024cambricon} uses Flash as a high-density memory to enable edge inference with 70B-parameter LLMs, combining near-flash compute with an NPU co-processor and efficient tiling.
Lincoln~\cite{sun2025lincoln} accelerates LLM inference on Compute-in-Flash by enabling an LPDDR interface for higher-bandwidth communication.

A concurrent work, KVNAND~\cite{deng2025kvnand}, aims to place both the KV cache and model weights on SSDs by optimizing KV placement to maximize device lifetime.
KVNAND’s own reliability evaluation estimates that continuous decoding of a 65B model at 3 tokens/s for 5 years generates ~143 TB of KV data, and the write volume scales approximately linearly with decoding throughput.
Recent SSD-based KV swapping systems make the endurance implication explicit: HiFC~\cite{jeong2025hifc} measures 1.98\,TiB/day of KV writes under a sustained long-context workload and shows that achieving multi-year lifetime requires confining KV traffic to a 200\,GiB pSLC region (6,000\,TiB TBW), whereas the same 1TB drive’s default TLC-mode endurance is only 750\,TBW.
Crucially, prior work has largely focused on (i) hardware support for flash-based compute-in-memory, or (ii) system-level scheduling and placement strategies to extend device lifetime.
However, there is limited work on algorithmic co-design that reshapes LLM inference itself to match Flash CIM constraints, especially limited floating-point support and write endurance.
Our work addresses this gap with two complementary contributions: (1) end-to-end integer-only execution for weights, KV cache, and nonlinear operations, enabling on-device computation without FP offloading, and (2) a dictionary-based KV cache compression scheme that minimizes KV write traffic and off-chip transfers while exploiting dense CIM arrays.

\subsection{Quantization}

LLM decoding has a large memory footprint due to the substantial weights and activations required for inference~\cite{gholami2024ai}.
Since decoding is often memory-bandwidth-bound~\cite{gholami2024ai}, reducing the size of weights and activations can improve throughput.

Quantization is a promising direction that reduces memory footprint and latency by representing weights and activations with lower-precision data types.
SmoothQuant~\cite{xiao2023smoothquant} observes that activations are harder to quantize than weights and shifts quantization difficulty from activations to weights by scaling activations down and weights up.
AWQ~\cite{lin2024awq} focuses on weight-only quantization and protects salient channels by scaling them to reduce quantization error.
OmniQuant~\cite{shao2024omniquant} similarly redistributes quantization difficulty between activations and weights through a differentiable calibration procedure, learning clipping thresholds instead of using standard min--max. SpinQuant~\cite{liu2024spinquant} takes a different approach by rotating weights and activations to mitigate outliers and improve quantizability.
Complementary directions replace expensive floating-point computation with table-based approximations: LUT-based methods precompute operator outputs and perform inference via table lookup~\cite{mo2025lut}, while NN-LUT~\cite{yu2022nn} learns neural approximations of nonlinear functions as a lookup table.

While quantization alone can provide speedups on GPUs, the overhead of dequantization and floating-point nonlinear/reduction operators can dominate on edge devices and alternative hardware~\cite{kim2021bert}.
I-BERT~\cite{kim2021bert} enables integer-only inference for Transformer-based BERT models by avoiding intermediate dequantization and approximating nonlinearities with integer polynomials.
I-LLM~\cite{hu2024llm} extends end-to-end integer inference to LLMs, using bit-shift-based approximations for nonlinear operators.

Despite these advances, many approaches still assume efficient floating-point support (e.g., for nonlinearities and reductions) or require dequantization at key operators, which is poorly matched to Flash/PIM controllers.
Motivated by SmoothQuant~\cite{xiao2023smoothquant} and I-BERT~\cite{kim2021bert}, our work targets end-to-end integer execution for modern LLM decoding by combining activation smoothing with finer-grained group-wise quantization and by redesigning integer nonlinear operators (e.g., Softmax and RMSNorm) to remain accurate under long-context, high-dynamic-range activations.

\subsection{KV Cache Compression}

Transformers~\cite{vaswani2017attention} are the backbone architecture of modern large language models (LLMs), but memory constraints arise as generation length grows.
During autoregressive decoding, attention reuses past keys and values by storing them in a key-value (KV) cache.
The KV cache grows linearly with sequence length, which makes memory capacity and bandwidth the dominant bottlenecks for generation~\cite{gholami2024ai}.

One line of work to address this memory constraint is KV-cache eviction, which reduces the memory footprint and thus saves memory bandwidth.
StreamingLLM~\cite{xiao2024efficient} maintains attention sink tokens and a local window, and H2O~\cite{zhang2023ho} leaves only ``heavy hitter'' important tokens.
SnapKV~\cite{li2024snapkv} evicts prompt tokens before decoding, whereas SpecKV~\cite{galim2026draftbased} uses an explicit drafter to guide which tokens to evict.

Another line of work is sparse attention, which maintains the full KV cache but changes the access pattern to save memory bandwidth.
MInference~\cite{jiang2024minference} uses three types of sparse attention patterns for the prompt during prefill, and Quest~\cite{tang2024quest} estimates token importance at page-level granularity of the KV cache using query vectors. SqueezeAttention~\cite{hooper2025squeezed} preprocesses the KV cache to create a hierarchical index, and retrieves the full KV cache only for important clusters during attention. MultipoleAttention~\cite{hooper2025multipole} further improves this by approximating the remaining unimportant KV-cache clusters with representative centroid values. 

Other interesting lines of work include quantizing the KV cache to lower precision~\cite{liu2024kivi}, or caching only quantized input activations and rematerializing KV values on-the-fly during inference~\cite{tomar2025xquant}.
While these approaches effectively reduce KV footprint and/or HBM bandwidth in GPU settings, they are primarily optimized around the GPU memory hierarchy.
Lexico~\cite{kim2025lexico} compresses the KV cache into sparse representations using an input-agnostic static dictionary.
Our KV-cache compression shares a similar idea with Lexico in maintaining a dictionary and sparse codes, but ours is carefully adapted for compute-in-memory constraints, whereas Lexico targets conventional GPU inference.
We also adopt an additional hierarchical sparsity scheme to further save read I/O.

%% file: _3_algorithm.tex
\section{Algorithm}

\subsection{Integer-Only LLM Quantization}

\subsubsection{Quantization Challenges for PIM}

Flash SSD controllers and near-flash processing engines provide limited support for high-precision floating-point arithmetic, making it challenging to execute Transformer decoding when key operators depend on FP nonlinear/reduction primitives~\cite{yu2024cambricon, ma2026challenges}.
To execute decoding near data, we aim to keep both linear layers and dominant nonlinear/reduction operators in the integer domain, minimizing dequantization and host round-trips.
Beyond device compatibility, integer execution also improves efficiency by reducing memory footprint and avoiding FP-heavy special-function logic that is costly in area and energy~\cite{kim2021bert}.

A direct obstacle is that many quantization pipelines~\cite{xiao2023smoothquant, shao2024omniquant, lin2024awq} are developed under GPU-centric assumptions: while weights and activations are quantized for storage or GEMM speedups, intermediate tensors often fall back to floating point for numerically sensitive operators such as Softmax and normalization.
Such FP fallbacks and dequantization complicate deployment on Flash/PIM controllers and can dominate end-to-end latency/energy when offloading is required.

We therefore target an integer-centric decoding pipeline tailored to Flash/PIM constraints.
We quantize weights and activations to INT8 (W8A8) for linear operators and implement key nonlinear/reduction operators using integer approximations, building on prior integer-only formulations~\cite{kim2021bert} but adapting them to modern LLM architectures and long-context regimes.
We find that maintaining accuracy under high-dynamic-range activations requires additional algorithmic adaptations, which we detail next.

\subsubsection{Activation Smoothing}

Naive INT8 quantization can incur substantial error due to activation outliers and high dynamic range.
We therefore adopt activation smoothing as in SmoothQuant~\cite{xiao2023smoothquant}, which rescales activations and compensates in the subsequent weights to make activations more quantization-friendly.
However, we find that the inputs and outputs to nonlinear operations are particularly sensitive and keeping these in high precision is vital to the overall fidelity of the model. 
To mitigate this, we use group-wise quantization along the hidden dimension, using per-group scale factors for both weights and activations to achieve finer quantization granularity.

\subsubsection{Nonlinear Operations}

We approximate nonlinear operations with second-order polynomial approximations, similar to the approximations used in I-BERT~\cite{kim2021bert}. Since modern LLMs such as LLaMA use SwiGLU (instead of GeLU), we also implement an integer approximation of SiLU (IntSiLU) using the same second-order $\exp(\cdot)$ approximation and use it to support SwiGLU in the integer pipeline.

Our group-wise quantization scheme poses a challenge for these polynomial approximations because both Softmax and RMSNorm require reduction across the hidden dimension. If activations are quantized with different scale factors across the hidden dimension, the reduction is no longer valid without dequantizing to full precision. However, if we do not quantize in groups along the hidden dimension, the quantization error becomes unacceptable, even though this is unnecessary for I-BERT~\cite{kim2021bert}. 
In our analysis, we find that longer context lengths and the larger activation dynamic range in LLMs lead to substantially higher quantization error compared to BERT.

To address this problem, we apply requantization to the input activations of Softmax and RMSNorm into token-wise INT16; that is, instead of using separate scale factors for each (\texttt{hidden dimension // group size}) group within a token, we use a single scale factor per token.
Additionally, since we employ a second-order approximation, requantizing to INT16 can cause intermediate values to overflow INT32.
To prevent this, we estimate overflow from the INT16 magnitude and right-shift the input before the approximation, then left-shift the output after the approximation for mathematical equivalence. For RMSNorm, we compute the required $1/\sqrt{\mathrm{var}}$ using an integer square-root routine based on Newton--Raphson iterations~\cite{akram2015newton}.
Finally, for both Softmax and RMSNorm, we avoid explicit division by using a fixed-point reciprocal (e.g., $2^{32}/x$) and implement normalization via 32-bit multiply-and-shift with appropriate bit-shifts during reductions to prevent overflow.

\subsection{Dictionary-Based KV Cache Compression}
\label{subsec:dictionary}

\subsubsection{Hardware Constraints and Design Rationale}
\label{subsubsec:hardware-and-design}
Flash cells have limited write endurance and can sustain only a finite number of program/erase cycles, making write-intensive workloads undesirable~\cite{ma2026challenges}.
At the same time, flash offers large, dense read-only capacity, allowing a substantial static dataset, such as a dictionary or codebook, to be stored persistently and reused without endurance concerns~\cite{yu2024cambricon, deng2025kvnand}.
These platforms also provide high internal (intra-array) read bandwidth and support simple near-data computations, which makes read-heavy operations particularly efficient~\cite{yu2024cambricon, deng2025kvnand, lee2025aif}.
However, a conventional KV cache is updated at every decoding step by appending new keys and values, inducing frequent writes and rendering a direct KV-cache placement in flash impractical~\cite{ma2026challenges}.

To align KV caching with these constraints, we represent the KV cache using a read-mostly formulation: a static dictionary is stored in flash, and per-token KV states are encoded as sparse codes (indices and coefficients). 
During attention, we operate directly on the codes via query--dictionary projections and lightweight gathers, while keeping only a small recent window in full precision.
Figure~\ref{fig:main-dictionary-compression} and Section~\ref{subsubsec:pipeline} summarize the end-to-end compression and attention pipeline.

\begin{figure}[t]
  \centering
  \includegraphics[width=0.7\columnwidth]{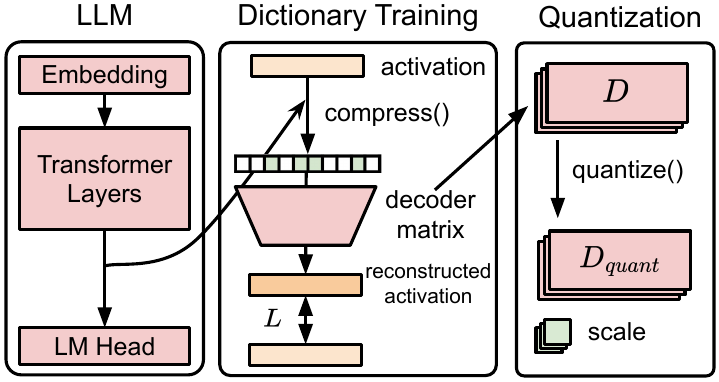}
  \caption{Hybrid sparse dictionary learning for KV compression. We extract attention activations, obtain Top-$K$ sparse codes using \texttt{compress()}, train the decoder as the dictionary, and per-tensor quantize the learned dictionary for flash CIM execution.}
  \label{fig:dictionary-learning}
  \vspace*{-3mm}
\end{figure}

\subsubsection{Learning the Overcomplete Dictionary}
\label{subsubsec:dictionary-training}
A key prerequisite for dictionary-based KV compression is a high-quality overcomplete dictionary that enables faithful sparse reconstructions of KV activations.
A natural baseline is classical sparse dictionary learning (e.g., K-SVD~\cite{aharon2006k} or MOD~\cite{engan1999method}), which alternates between a sparse-coding step and a dictionary-update step.
However, applying these algorithms at the scale of LLM activations is often prohibitively expensive and difficult to scale, making it challenging to obtain a robust dictionary over a large language corpus.
An alternative is to train a sparse autoencoder~\cite{bricken2023towards, cunningham2023sparse, gao2025scaling}, where an encoder produces sparse coefficients and a decoder serves as an overcomplete dictionary.
In our preliminary experiments, we find that encoder--decoder co-evolution can be unstable, leading to suboptimal reconstruction quality under tight sparsity budgets.

Therefore, we adopt a training-based hybrid approach: we replace the encoder with a non-parameterized sparse-coding procedure and optimize only the decoder (dictionary) parameters, building on open-source sparse autoencoder (SAE) implementations~\cite{gao2025scaling}.
We extract key and value activations from each attention layer, buffer them in a large shuffled pool to reduce temporal correlations, and train a separate dictionary per layer and per stream (key/value).
Sparse codes are produced by our projection-based \texttt{compress()} procedure (Algorithm~\ref{alg:compress}), which we detail in Section~\ref{subsubsec:sparse-compression}; the decoder matrix is then trained to reconstruct the original activations by minimizing reconstruction MSE.
After training, we apply per-tensor quantization to each dictionary using a single scale factor, enabling efficient storage and compute on flash CIM devices.
We summarize the full training pipeline in Figure~\ref{fig:dictionary-learning}.

\begin{algorithm}[t]
\caption{Projection-Based Sparse Coding (Compression)}
\label{alg:compress}
\begin{algorithmic}[1]
\Require Dictionary $D \in \mathbb{R}^{m \times n}$ with column-normalized atoms $\{\mathbf{d}_j\}_{j=1}^n$, input vector $\mathbf{x} \in \mathbb{R}^{m}$, sparsity budget $K$
\Ensure Support indices $I$ and coefficients $\boldsymbol{\gamma}\in\mathbb{R}^{n}$ such that $\mathbf{x}\approx D\boldsymbol{\gamma}$ with $\|\boldsymbol{\gamma}\|_0 \le K$
\State $\mathbf{r} \gets \mathbf{x}$ \Comment{residual}
\State $I \gets \emptyset$, $\boldsymbol{\gamma} \gets \mathbf{0}$
\For{$t = 1$ to $K$}
    \State $\mathbf{c} \gets D^{\mathsf{T}}\mathbf{r}$ \Comment{correlations (linear projection)}
    \State $j^\star \gets \arg\max_{j \notin I} |c_j|$ \Comment{avoid duplicates}
    \State $I \gets I \cup \{j^\star\}$
    \State $\gamma_{j^\star} \gets \gamma_{j^\star} + c_{j^\star}$ \Comment{use correlation as coefficient}
    \State $\mathbf{r} \gets \mathbf{r} - c_{j^\star}\mathbf{d}_{j^\star}$ \Comment{subtract selected atom contribution}
\EndFor
\State \Return $I, \boldsymbol{\gamma}$
\end{algorithmic}
\end{algorithm}

\subsubsection{Projection-Based Sparse Coding}
\label{subsubsec:sparse-compression}

During decoding, given a pretrained dictionary $D$, we compress each full-precision key/value vector $\mathbf{x}$ into a sparse representation by selecting a small set of dictionary atoms and storing only their indices and coefficients. We use the same projection-based \texttt{compress()} operator both to generate training codes (Section~\ref{subsubsec:dictionary-training}) and to encode KV states at inference time, avoiding train--inference mismatch.

A standard approach is Orthogonal Matching Pursuit (OMP)~\cite{pati1993orthogonal}, which iteratively expands a support set and re-solves a least-squares refit over the selected atoms at every step; this class of OMP-style coding is also used in prior dictionary-based KV compression (e.g., Lexico~\cite{kim2025lexico}).
While effective, this refit requires repeated least-squares solves (e.g., \texttt{lstsq}), which are computationally expensive and poorly matched to compute-in-memory (CIM) primitives.

We instead use a projection-based greedy pursuit that eliminates least-squares refitting and relies only on linear projections and residual updates (Algorithm~\ref{alg:compress}).
Starting from the residual $\mathbf{r}\leftarrow\mathbf{x}$, each iteration (i) computes correlations $\mathbf{c}=D^{\mathsf{T}}\mathbf{r}$, (ii) selects the atom whose correlation magnitude is largest, $j^\star=\arg\max_{j\notin I}|c_j|$, (iii) appends $j^\star$ to the support set $I$, (iv) uses the correlation value directly as the coefficient for that atom, and (v) updates the residual by subtracting the selected contribution, $\mathbf{r}\leftarrow \mathbf{r}-c_{j^\star}\mathbf{d}_{j^\star}$.
By construction, this procedure performs only matrix--vector projections and scaled vector updates, making it substantially more hardware-friendly than OMP on CIM devices.
Overall, the projection-based method provides a favorable accuracy--efficiency trade-off for our flash CIM setting.

\begin{figure}[t]
  \centering
  \includegraphics[width=0.7\columnwidth]{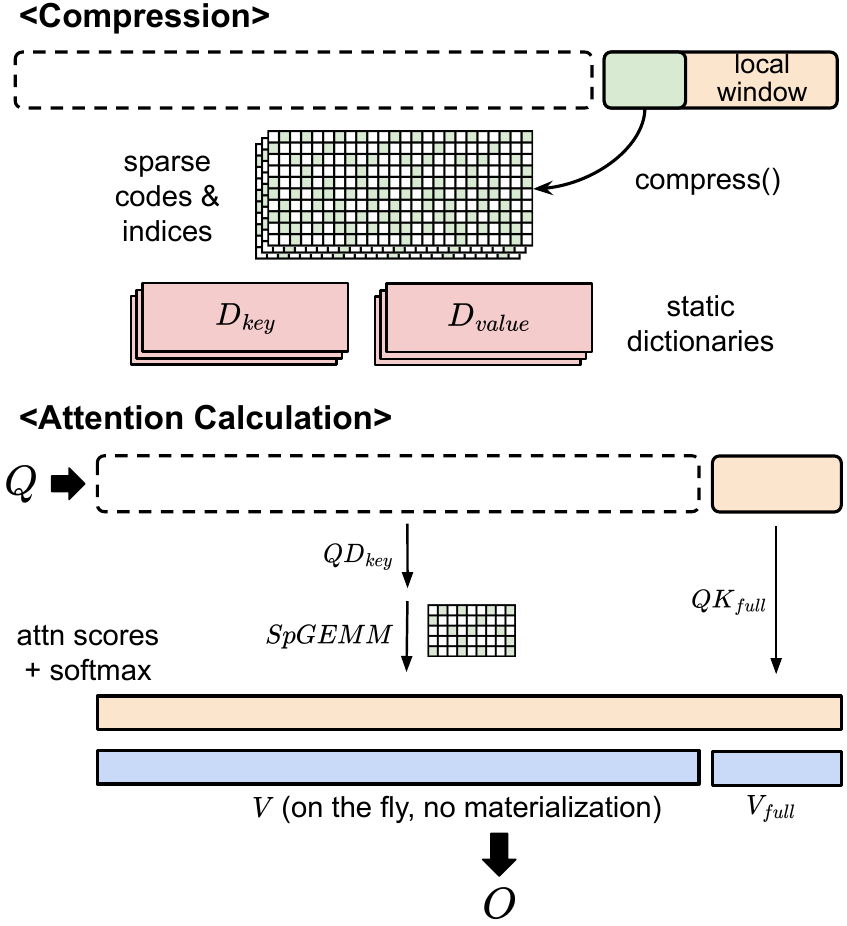}
  \caption{End-to-end dictionary-based KV cache compression for flash CIM. Top: KV compression into sparse codes with static dictionaries. The dotted region shows the KV cache that are already approximated as static dictionaries, and the green region shows the local buffer of KV cache tokens that are being compressed as sparse codes and indices. Bottom: on-the-fly attention over compressed KV (no materialization), producing output O. The dotted region shows the KV cache which are being stored as sparse codes and indices (for which we compute a sparse GEMM operation).
  }
  \label{fig:main-dictionary-compression}
  \vspace*{-3mm}
\end{figure}

\begin{figure*}[t]
  \centering
  \includegraphics[width=\textwidth]{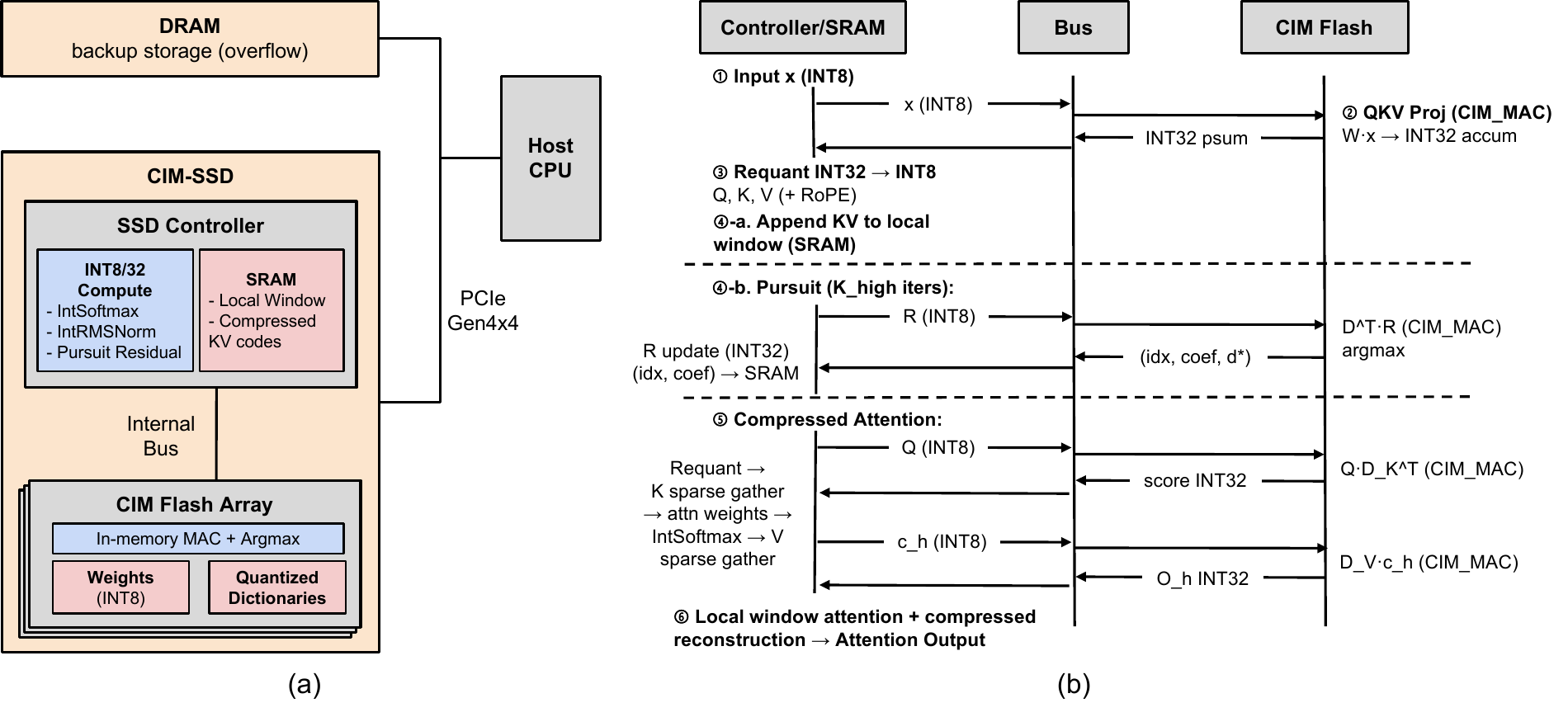}
  \caption{
  System overview for analytical modeling.
  (a) Modeled CIM-SSD platform: a host CPU with DRAM backing and a PCIe-attached CIM-SSD comprising an SSD controller with SRAM and a CIM flash array.
  (b) Modeled decode pipeline and data movement across controller/SRAM, bus, and CIM flash.
  }
  \label{fig:system-modeling}
  \vspace*{-3mm}
\end{figure*}

\subsubsection{Overall Pipeline.}
\label{subsubsec:pipeline}
Figure~\ref{fig:main-dictionary-compression} illustrates our runtime pipeline for dictionary-based KV cache compression and attention during LLM decoding.
We maintain a two-part KV cache: (i) a long \emph{compressed context} represented by sparse codes over a pretrained dictionary, and (ii) a recent \emph{local window} stored in full precision to preserve fidelity for the most recent tokens.
During generation, newly generated KV entries are first appended to the local window (yellow blob).
When the window grows beyond its target size (we use a fixed local window size $W$ in this work), we \emph{batch-compress} the oldest tokens in the window (green blob) in fixed-size chunks of $U$ tokens (the \emph{compression unit}), convert their keys and values into sparse codes using the pretrained dictionary, append these codes to the compressed context (dotted region), and evict the corresponding dense KV tensors from the raw cache. 

For the compressed context, we avoid explicit materialization of keys and values.
We first project the query onto the dictionary ($Q D^{\mathsf{T}}$) and use the sparse codes to assemble approximate attention logits $QK^{\mathsf{T}}$.
In parallel, we compute exact attention over the dense local window.
We then compute the attention output by aggregating values directly from the value dictionary using the stored sparse coefficients, and combine contributions from the compressed context and the local window under a single softmax normalization.
To support this pipeline efficiently, the runtime stores only sparse codes for the compressed region and a bounded dense buffer for the local window.

\subsubsection{Query-Aware Hierarchical Sparsity}
\label{subsubsec:hierarcical-sparsity}
To further reduce compute and read I/O, we allocate reconstruction budget non-uniformly across tokens in a query-aware manner, inspired by prior work on query-aware sparsification and hierarchical attention~\cite{hooper2025multipole}.
Our key idea is a two-stage refinement scheme: we first construct a coarse reconstruction for all tokens using a small budget $K_{\mathrm{low}}$, and then selectively refine only the most relevant tokens up to a larger budget $K_{\mathrm{high}}$.

Concretely, we form a coarse key approximation for each token $i$ using only its first $K_{\mathrm{low}}$ dictionary atoms and compute a proxy attention logit:
\begin{equation}
\mathrm{Score}(i)
= Q \cdot \Big(\sum_{j=1}^{K_{\mathrm{low}}} \gamma_{i,j}\,\mathbf{d}_{\mathrm{idx}_{i,j}}\Big).
\end{equation}
We then select the top-$N\%$ tokens by $\mathrm{Score}(i)$ and refine them by increasing the reconstruction budget from $K_{\mathrm{low}}$ to $K_{\mathrm{high}}$.
We compute attention under a single numerically stable softmax over this mixed representation (top tokens at $K_{\mathrm{high}}$, others at $K_{\mathrm{low}}$), concentrating dictionary reads and near-data compute on the most relevant tokens.

\subsubsection{Why Flash CIM?}
\label{subsubsec:why-CIM}

Our dictionary-based KV representation is well matched to flash-based compute-in-memory (CIM), but less attractive on conventional accelerators such as GPUs, as they do not support compute-in-memory.
The method requires repeated projections against a large dictionary (e.g., $Q D^{\mathsf{T}}$) and gather-style aggregation from sparse codes.
On conventional accelerators, this often means repeatedly streaming the dictionary (and related intermediates) from HBM, increasing memory traffic and potentially dominating runtime.
Moreover, LLM decoding is frequently memory-bandwidth-bound even without compression; the additional dictionary reads and compute can offset the benefits of a more compact KV representation, yielding limited gains or slowdowns.
In contrast, flash CIM can keep the dictionary resident in flash and execute projection/gather near data, exploiting high internal read bandwidth while reducing off-chip transfers.

That said, dictionary-based KV coding can still be useful on conventional accelerators when the KV cache no longer fits in device memory (e.g., very long contexts or large batches), by storing sparse codes on the accelerator and reducing host--device offload bandwidth.



%% file: _4_system_modeling.tex
\section{System Modeling}
\label{sec:system-modeling}

\paragraph{Overview and metrics.}
We implement an analytical Python model to compare a PCIe-attached inference NPU baseline against our CIM-SSD (\emph{LLM-in-a-Flash}).
The model targets absolute estimates grounded in published hardware data (see Appendix \ref{sec:appendix-hw-data}) and is designed to capture relative trends with respect to context length, controller SRAM capacity, and pursuit placement.
We focus on decoding behavior and report latency (throughput) and total per-token energy, with a breakdown that highlights KV write and read costs.

\paragraph{Systems and workload.}  
We compare four configurations spanning two NPU baselines and two CIM-SSD variants.
The first baseline \textbf{B1} is an inference NPU with DDR5 DRAM that stores both model weights and the full FP16 KV cache.
A second baseline \textbf{B2} uses the same NPU+DRAM hardware but applies our dictionary-based KV cache compression, storing dictionaries and compressed codes in DRAM and performing compression/reconstruction on the NPU.

The proposed CIM-SSD system \textbf{S1} integrates an SSD controller (INT8/INT32) with on-controller SRAM, a CIM flash array that stores stationary weights and per-layer dictionaries, and DRAM as a backing store (Figure~\ref{fig:system-modeling}(a)).
We assume PCIe Gen4$\times$4 (8\,GB/s) for the host--SSD link and a 16\,GB/s internal bus between the controller and CIM array.
Additionally, we explore a second configuration, \textbf{S2}, which  accounts for limited on-controller SRAM capacity, placing compressed codes on DRAM and loading this to the CPU during reconstruction (while still applying pursuit on CIM).

All configurations model Llama~3.1--8B with GQA ($L{=}32$, $H_Q{=}32$, $H_{KV}{=}8$, $d{=}128$), keeping $H_Q$ (query-side compute) separate from $H_{KV}$ (stored KV state) so that operations scaling with query heads versus stored KV heads are modeled distinctly.
We focus on autoregressive decoding (batch size 1), where KV-cache access dominates.



\paragraph{Cost primitives and overlap.}
We decompose each decode step into compute and data movement.
Each engine has peak throughput $P$ (TOPS). For $N_{\text{op}}$ operations,
\begin{equation}
t_{\text{comp}}=\frac{N_{\text{op}}}{P}, \qquad
E_{\text{comp}}=N_{\text{op}} \cdot e_{\text{op}} + \left\lceil \frac{N_{\text{op}}}{S} \right\rceil \cdot e_{\text{ovhd}},
\end{equation}
where $S$ is SIMD width and $e_{\text{ovhd}}$ is a per-SIMD-group instruction overhead.
Transfers follow a bandwidth model with optional fixed protocol terms for interconnects:
\begin{equation}
t_{\text{xfer}}= t_\ell + t_o + \frac{\text{Bytes}}{B}, \qquad
E_{\text{xfer}} = 8 \cdot \text{Bytes} \cdot e_b,
\end{equation}
where $t_\ell$ is the latency before the transfer starts (link latency), and $t_o$ is the fixed overhead associated with initiating the transfer, with $(t_\ell,t_o)\!>\!0$ for PCIe/internal buses and $(t_\ell,t_o)\!=\!0$ for DRAM/flash, and where $B$ is the bandwidth. Primitive energies follow Horowitz-style baselines~\cite{horowitz20141} with 7\,nm-class calibration guided by TPU-era references~\cite{jouppi2021ten}; CIM periphery is treated as an older-node process.

Within a stage, we model overlap between independent compute and communication engines using
\begin{equation}
\texttt{overlap}(t_1,t_2)=\max(t_1,t_2)+(1-\rho)\min(t_1,t_2),
\end{equation}
with $\rho{=}0.8$ for the NPU baseline and $\rho{=}0.9$ for CIM-SSD, applied only when concurrency is plausible (e.g., CIM matmul with internal-bus transfers).
Across stages of a decode step (QKV projection $\rightarrow$ KV write $\rightarrow$ KV read/attention $\rightarrow$ FFN), execution is sequential due to shared engine/channel contention.
Pursuit iterations are also sequential ($\rho{=}0$) due to $\mathbf{r}^{(i+1)}=\mathbf{r}^{(i)}-\alpha_i\mathbf{d}_{j_i}$.

\paragraph{Layer modeling.}

Each decode layer follows a standard transformer decode pipeline (QKV projection, RoPE, KV-cache update, attention, output projection, residual paths, SwiGLU FFN, and RMSNorm). Figure~\ref{fig:system-modeling}(b) illustrates the CIM-SSD dataflow.

\textbf{Linear layers} (QKV, output, and FFN projections) are modeled as GEMV at batch size~1.
In the baseline, the NPU streams FP16 weights from DRAM every token, making these projections primarily memory-bandwidth bound.
In CIM-SSD (Figure~\ref{fig:system-modeling}(b), \circled{1}--\circled{3}), weights remain stationary in flash: the controller broadcasts the INT8 activation $\mathbf{x}$ over the internal bus, the CIM array performs in-place INT8 MAC, and INT32 partial sums return to the controller for requantization back to INT8.
CIM compute and internal-bus transfers are overlapped within the linear stage.

\textbf{Nonlinear operations} (RMSNorm, softmax, SiLU, RoPE, residual add, and post-Swish multiply) run on the NPU in FP32 for the baseline.
On CIM-SSD, they execute on the SSD controller using integer-only approximations (INT8/INT32 + requantization), e.g., polynomial/shift-based $\exp$ for IntSoftmax and integer sqrt/divide for IntRMSNorm, without relying on floating-point units.

\paragraph{KV cache write path.}

Each new token appends its INT8 key and value vectors to a \emph{local window} in controller SRAM (Figure~\ref{fig:system-modeling}(b), \circled{4}-a).
Periodically, the oldest entries are batch-compressed using pursuit (\circled{4}-b): the controller sends the current residual $\mathbf{r}$ (INT8) to the CIM array over the internal bus; the CIM array computes $D^\top\!\mathbf{r}$ in-place and returns the selected atom and its corresponding index/coefficients; the controller then updates the residual in INT32 and records the sparse code.
After a fixed number of pursuit iterations, each token is represented as a compact set of (index, coefficient) pairs stored in SRAM, while all intermediate state (residual buffers and partial codes) remains on-controller throughout the loop. In the baseline (\textbf{B1}), KV vectors are written in FP16 directly to DRAM.
In \textbf{B2}, the same compression runs on the NPU, but dictionaries and codes reside in DRAM, introducing additional DRAM transfers during encoding.

\paragraph{KV cache read path.}

The read path combines two regions, an uncompressed \emph{local window} and a \emph{compressed context}, with scenario-specific data movement (Figure~\ref{fig:system-modeling}(b), \circled{5}--\circled{6}). \textbf{B1} reads the full FP16 KV cache from DRAM and runs standard attention on the NPU ($QK^\top$, softmax, and $\text{Score}\cdot V$). \textbf{B2} reads compressed codes \emph{and} per-layer dictionaries from DRAM, reconstructs KV on the NPU, and then runs standard attention. The repeated dictionary transfer becomes a fixed per-layer cost during decoding.

\textbf{S1 (CIM-SSD)} avoids both full-KV transfer and dictionary loading. For the \emph{local window}, the controller reads INT8 KV from SRAM and computes attention locally. For the \emph{compressed context}, attention operates in the dictionary-score domain without materializing full KV: the CIM array computes $Q D_K^\top$, the controller combines the resulting scores with sparse codes to form attention weights and code aggregates, and the CIM array applies $D_V$ to produce the value-side contribution. The local-window and compressed-context outputs are then combined.
\textbf{S2} considers the memory constraints of the local buffer on the SSD by storing codes in DRAM and fetching them to the CPU when performing reconstruction.

\paragraph{SRAM budget and overflow.}


Controller SRAM must hold (i) an uncompressed local window (recent INT8 KV), and (ii) transient pursuit state (residual buffers, partial outputs, and control metadata) across all layers. 
For \textbf{S1}, it also must store compressed KV history codes (growing with sequence length).
Dictionaries are \emph{not} stored in SRAM; they reside permanently in CIM flash.
The dominant term is the compressed-history codes, which scale linearly with context length. 
In \textbf{S1}, the SRAM budget is sized to keep both the local window and the history codes on-controller for typical sequence lengths.
With \textbf{S2}, the coefficients are stored in DRAM.

%% file: _5_results.tex
\section{Results}

We evaluate our approach on Llama-3.1-8B and Qwen-2.5-7B.
We first ablate each component of our integer-only quantization pipeline (Section~\ref{subsec:results-intonly}) and dictionary-based KV cache compression (Section~\ref{subsec:results-kv}), and then present combined downstream task results (Section~\ref{subsec:results-main}).
Finally, we report system-level latency and energy projections from our analytical model (Section~\ref{subsec:results-system}).

\subsection{Experimental Setup.}
For integer-only quantization, we calibrate smoothing parameters and quantization scales using the Pile validation set following SmoothQuant ~\cite{xiao2023smoothquant}, and use group size of 128 for group-wise quantization.
For dictionary training, we extract key and value activations from Pile calibration sequences, train a separate overcomplete dictionary ($M{=}32768$ atoms) per layer and per key/value, and apply per-tensor INT8 quantization to the learned dictionaries.
We use a local window of $W{=}128$ tokens, and when this is full we compress the oldest $U=32$ tokens from the window.
We use a sparsity budget of $K_{\mathrm{high}}{=}16$ with $K_{\mathrm{low}}{=}3$ for the hierarchical policy.

\subsection{Integer-Only LLM Inference}
\label{subsec:results-intonly}

We evaluate the impact of our integer-only quantization pipeline on language modeling quality, reporting perplexity on WikiText-2~\cite{merity2016pointer} for Llama-3.1-8B.
Table~\ref{tab:intonly-ablation} ablates each design decision cumulatively.

Starting from the FP16 baseline, naive per-tensor W8A8 quantization incurs substantial perplexity degradation due to activation outliers.
Adding SmoothQuant \cite{xiao2023smoothquant} significantly recovers quality by redistributing quantization difficulty from activations to weights.
Group-wise quantization along the hidden dimension further improves accuracy by providing finer-grained scale factors for both weights and activations.
Replacing floating-point nonlinear operations (Softmax, RMSNorm, SiLU) with our integer polynomial approximations introduces a modest perplexity increase, as the second-order approximations trade off small numerical fidelity for full integer-domain execution.
This degradation is mitigated by applying INT16 requantization at Softmax and RMSNorm inputs.
Our final end-to-end integer-only pipeline achieves 7.5803 perplexity, within 0.0349 of the FP16 baseline, while eliminating all floating-point computation from the critical path.

\begin{table}[!htbp]
    \caption{
    Cumulative ablation of integer-only quantization on Llama-3.1-8B.
    Each row adds one component to the previous configuration.
    We report WikiText-2 perplexity ($\downarrow$).
    * Applying integer nonlinear approximations naively results in the model activations diverging and resulting in NaN.
    }\label{tab:intonly-ablation}
    \centering
    \small
    \begin{tabular}{lc}
    \toprule
    \textbf{Configuration} & \textbf{WikiText-2 PPL ($\downarrow$)} \\
    \midrule
    FP16 baseline                          & 7.5454 \\
    + Per-tensor W8A8                      & 7.6525 \\
    + SmoothQuant and group-wise quant  ($g{=}128$)                          & 7.5511 \\
    + Integer nonlinear approx.            &  NaN* \\
    + INT16 requantization                 &  7.5803 \\
    \bottomrule
    \end{tabular}
    \end{table}

\subsection{Dictionary-Based KV Cache Compression}
\label{subsec:results-kv}

We evaluate our dictionary-based KV cache compression on LongBench, a benchmark suite for long-context understanding that covers single- and multi-document QA, summarization, few-shot learning, and code completion tasks.
We report the average score across all LongBench tasks.
Table~\ref{tab:kv-ablation} ablates each component of our compression pipeline, showing the accuracy versus KV cache size trade-off with 1) the baseline dictionary compression method, 2) our dictionary compression method, 3) INT8 quantization of the learned dictionaries, and 4) our query-aware hierarchical sparsity scheme.

Dictionary compression with OMP-style coding at $K{=}16$ atoms per token provides a strong baseline, achieving a 4$\times$ reduction in KV cache size (in terms of dynamic memory requirements) with minimal accuracy degradation.
Replacing OMP with our projection-based sparse coding (see Algorithm~\ref{alg:compress}) yields comparable accuracy while eliminating the costly least-squares operations, making the compression step substantially more hardware-friendly for CIM execution.
Applying per-tensor INT8 quantization to the learned dictionaries introduces minimal additional quality loss, as the overcomplete dictionary is robust to moderate quantization noise.
Finally, our query-aware hierarchical sparsity scheme uses $K_{\mathrm{low}}{=}3$ atoms for coarse scoring and refines the top-10\% of tokens to $K_{\mathrm{high}}{=}16$, further reducing read I/O with only a small accuracy trade-off.
The combined pipeline reduces the effective KV cache footprint by 15$\times$ without degrading the LongBench score (see Table \ref{tab:kv-ablation}).

\begin{table}[!htbp]
    \caption{
    Cumulative ablation of dictionary-based KV cache compression on LongBench (Llama-3.1-8B).
    Each row adds one component to the previous configuration.
    We report per-task scores and the effective KV cache compression ratio (according to the amount of dynamic memory that we need to load at each decoding step, ignoring the footprint of the dictionary).
    }\label{tab:kv-ablation}
    \centering
    \scriptsize
    \setlength{\tabcolsep}{3.5pt}
    \begin{tabular}{l|ccccc|c|c}
    \toprule
    \multirow{2}{*}{\textbf{Configuration}} &
    \textbf{multifield} & \textbf{2wiki} & \textbf{gov\_} & &  & & \textbf{KV} \\
    
    & \textbf{qa-en} & \textbf{mqa} & \textbf{report} & \textbf{trec} & \textbf{lcc} & \textbf{Avg.} & \textbf{Compr.} \\
    \midrule
    Full FP16 KV cache
        & 32.79 & 13.99 & 30.2 & 73.0 & 70.35 & 44.07 & $1\times$ \\
    \hc Sparse coding (K=16)
        & 37.37 & 13.71  & 29.22 & 73.5 & 69.25 & 44.61 & 4$\times$ \\
    \hd + Hierarchical
        & 36.06 & 13.75 & 29.71 & 73.5 & 69.76 & 44.56 & 15$\times$ \\
    \bottomrule
    \end{tabular}
    \end{table}

\subsection{End-to-End Results}
\label{subsec:results-main}

Table~\ref{tab:main-results} presents the end-to-end evaluation of our combined pipeline (integer-only quantization + dictionary-based KV cache compression) on LongBench across two models.
Our combined pipeline retains 99\% of the FP16 full-KV baseline score for both Llama-3.1-8B and Qwen-2.5-7B. Notably, applying our joint integer-only quantization and sparse dictionary KV cache compression method degrades performance by less than 1 point on average, while providing nearly 15$\times$ KV compression compared to the floating point version.

\begin{table}[!htbp]
    \caption{
    End-to-end LongBench evaluation of our combined pipeline (integer-only quantization + dictionary-based KV cache compression) on Llama-3.1-8B and Qwen-2.5-7B. We report per-task scores and the effective KV cache compression ratio (according to the amount of dynamic memory that we need to load at each decoding step, ignoring the footprint of the dictionary).
    }\label{tab:main-results}
    \centering
    \scriptsize
    \setlength{\tabcolsep}{3.5pt}
    \begin{tabular}{l|ccccc|c|c}
    \toprule
    \multirow{2}{*}{\textbf{Configuration}} &
    \textbf{multifield} & \textbf{2wiki} & \textbf{gov\_} & &  & & \textbf{KV} \\
    
    & \textbf{qa-en} & \textbf{mqa} & \textbf{report} & \textbf{trec} & \textbf{lcc} & \textbf{Avg.} & \textbf{Compr.} \\
    \midrule
    \multicolumn{8}{c}{\textbf{Llama-3.1-8B}} \\
    \midrule
    Full FP16 KV cache
        & 32.79 & 13.99 & 30.2 & 73.0 & 70.35 & 44.07 & $1\times$ \\
    \hd Ours 
        & 31.17 & 13.95 & 30.41 & 73.0 & 69.71 & 43.65 & 15$\times$ \\
    \midrule
    \multicolumn{8}{c}{\textbf{Qwen-2.5-7B}} \\
    \midrule
    Full FP16 KV cache
        & 35.36 & 15.98 &  32.2 & 72.0 &  66.68 & 44.44 & $1\times$ \\
    \hd Ours
        & 40.01 & 19.52 &  29.64 & 72.5 & 59.48 & 44.23 & 15$\times$ \\
    \bottomrule
    \end{tabular}
    \end{table}


\subsection{System-Level Modeling}
\label{subsec:results-system}

We use the analytical model described in Section~\ref{sec:system-modeling} to project system-level latency and energy for our CIM-SSD design, modeling Llama-3.1-8B at batch size 1 across varying context lengths.
We report time-between-tokens (TBT) for decode latency and total energy per decode step.
We compare four configurations spanning two baselines and two CIM-SSD variants:

\begin{itemize}
    \item \textbf{B1 -- NPU, no compression.}
    NPU + DDR5 DRAM, FP16 inference.
    Full KV cache written to and read from DRAM each decode step.

    \item \textbf{B2 -- NPU + dictionary compression.}
    NPU + DDR5 DRAM, FP16 inference with dictionary-based KV compression.
    Dictionary and compressed codes are stored in DRAM, with pursuit being applied on the CPU.

    \item \textbf{S1 -- CIM-SSD, coefficients on controller.}
    CIM-SSD, INT8 inference, CIM-resident dictionary.
    SRAM stores the local window, compressed codes, and temporaries for the pursuit algorithm (no overflow).
    \item \textbf{S2 -- CIM-SSD, coefficients on DRAM.}
    CIM-SSD + DDR5 DRAM, INT8 inference, CIM-resident dictionary.
    Pursuit runs on the controller (same as S1), but compressed codes are written to DRAM.
    During attention, codes are read from DRAM to the CPU, and intermediate results are synchronized between CPU and SSD via PCIe.
\end{itemize}

Table~\ref{tab:sys-combined} summarizes all results for all four configurations for short (1K)/long (256K) context length inference (additional results for 64K context length are provided in Appendix \ref{sec:appendix-full-system-results}).
Note that the headline savings reported in the abstract and contributions ($3.1\times$/$2.7\times$ latency/energy at 1K and $4.4\times$/$6.8\times$ latency/energy at 256K) correspond to configuration \textbf{S2}, which we treat as our default serving configuration since it supports both short and long context lengths.

\begin{table}[!htbp]
    \caption{
    System-level decode latency (TBT, ms) and energy (J) across context lengths for Llama-3.1-8B (batch size 1).
    B1/B2 are NPU+DRAM baselines; S1/S2 are CIM-SSD variants.
    Parenthesized values show speedup ($\uparrow$) / energy reduction ($\uparrow$) relative to B1.
    }\label{tab:sys-combined}
    \vspace*{-2mm}
    \centering
    \scriptsize
    \setlength{\tabcolsep}{3pt}
    \begin{tabular}{ll|cc|cc}
    \toprule
    & & \multicolumn{2}{c|}{\textbf{1K}} & \multicolumn{2}{c}{\textbf{256K}} \\
    & \textbf{Configuration} & TBT & Energy & TBT & Energy \\
    \midrule
    B1 & NPU, no compr.             & 183.7 & 101.8 & 631.5 & 354.5 \\
    B2 & NPU + dict.\ compr.        & 190.4 (0.96$\times$) & 108.2 (0.94$\times$) & 217.4 (2.91$\times$) & 135.1 (2.62$\times$) \\
    \midrule
    \hd S1 & CIM-SSD, coeff.\ on ctrl.  & 54.1 (3.40$\times$)  & 37.4 (2.72$\times$)  & 208.4 (3.03$\times$) & 40.9 (8.66$\times$)  \\
    \hd S2 & CIM-SSD, coeff.\ on DRAM   & 59.9 (3.07$\times$)  & 37.8 (2.69$\times$)  & 143.5 (4.40$\times$) & 52.5 (6.75$\times$)  \\
    \bottomrule
    \end{tabular}
    \end{table}

\paragraph{End-to-end comparison (B1 vs.\ S1).}
Comparing the uncompressed NPU baseline against our CIM-SSD pipeline isolates the combined benefit of integer-only execution, CIM-resident weights, and dictionary-based KV compression.
For short context lengths (1K tokens), S1 is $3.4\times$ faster and $2.7\times$ more energy-efficient than B1, as CIM-resident weights eliminate the dominant DRAM weight-streaming cost.
For longer context lengths (256K tokens), the energy gap widens to $8.7\times$ while maintaining a $3.0\times$ latency advantage, with energy savings growing as KV cache I/O dominates.

\paragraph{Value of compression on NPU (B1 vs.\ B2).}
Adding dictionary compression to the NPU baseline provides reduced benefit because the dictionary itself must be loaded from DRAM each decode step, introducing a fixed data-movement cost that offsets the savings from compact KV codes.
At 1K context, B2 is slightly \emph{slower} than B1 ($0.96\times$) and less energy-efficient ($0.94\times$).
The gains only materialize at long sequences: at 256K, B2 is $2.9\times$ faster and $2.6\times$ more energy-efficient, demonstrating that while our compression algorithm can reduce KV transfers even on non-CIM hardware, it is particularly well-suited to CIM.

\paragraph{CIM advantage for compression (B2 vs.\ S1).}
The CIM-SSD architecture dramatically reduces the overhead from loading the dictionary: dictionaries reside stationary in the CIM flash array and are accessed via high-bandwidth internal channels rather than DRAM.
At 1K context, S1 is already $3.5\times$ faster than B2 and $2.9\times$ more energy-efficient, since CIM avoids both the dictionary load and the weight-streaming costs.
At 256K, S1 and B2 achieve comparable latency, but S1 uses $3.3\times$ less energy because it avoids streaming the dictionary from DRAM.

\paragraph{Coefficient placement trade-off (S1 vs.\ S2).}
S1 and S2 represent two strategies for where compressed codes reside: on-controller SRAM (S1) versus host DRAM (S2), with S2 being preferred for long contexts where SRAM capacity is exceeded.
At 1K context, S1 is slightly faster because all codes fit in SRAM and no PCIe synchronization is needed.
However, at longer contexts S2 is faster than S1 (with a $1.5\times$ latency advantage for S2 at 256K context length).
This occurs because the controller has limited computational resources (50 GFLOPS), whereas the CPU has higher throughput (2 TFLOPS) and bandwidth (76.8 GB/s DDR5), leading to a $1.5\times$ latency advantage for S2.
However, S1 remains more energy-efficient across all context lengths since on-controller reads avoid DRAM transfer energy.
In practice, S1 is suitable when all codes fit in SRAM, while S2 provides a more generalizable solution for long contexts; we hence reference the S2 results as the assumed configuration when reporting summarized data.

%% file: _6_conclusion.tex
\section{Conclusion}

With LLM inference becoming a predominant workload for enabling state-of-the-art performance across a range of downstream applications, the memory requirements of LLMs has become a critical bottleneck.
Compute-in-Flash technologies have high potential to address the memory bottleneck by eliminating off-chip memory transfers; however, these devices have limited write endurance and support for high-precision floating point computation, as well as low off-chip bandwidth relative to HBM.
In this work, we exploit Compute-in-Flash to address the memory bottleneck with LLM inference. 
We co-design LLM inference to exploit the benefits of in-memory computation by adapting the algorithm to only use efficient integer computations, and adapting the KV cache management to use read-only information to exploit the high internal read bandwidth.
These adaptations enable \textbf{3.1}$\times$/\textbf{2.7}$\times$ latency/energy savings for short context lengths and \textbf{4.4}$\times$/\textbf{6.8}$\times$ latency/energy savings for long context lengths.
Our approach demonstrates algorithmic solutions to the constraints of Compute-in-Flash, thereby enabling efficient inference and reducing the core memory bottleneck with LLM inference.

\section{Limitations}

One limitation of our study is that it focuses on the generation phase of LLM inference.
The compute-bound prefill phase is more challenging to map efficiently to compute-in-memory platforms (as these platforms typically have lower total peak compute performance than CPU/GPU platforms).
Exploring how to map both the prefill and decode phase efficiently is important future work.
A second limitation is that the Value dictionary is access both transposed (during pursuit) and non-transposed (when computing attention).
To realize the energy and latency benefits of compute-in-memory for both pursuit and computing attention, we therefore assume that 2 copies of the Value dictionary are stored (one transposed and one non-transposed).

%% file: _7_appendix.tex
\newpage

\section{System Modeling Parameters}
\label{sec:appendix-hw-data}

Tables~\ref{tab:hw-compute-energy}-\ref{tab:hw-throughput} summarize the hardware parameters used in our analytical system model (Section~\ref{sec:system-modeling}).

\paragraph{Compute energy.}
Table~\ref{tab:hw-compute-energy} lists per-operation energy for CPU and SSD-CIM.
CPU operation energies are from \cite{jouppi2021ten}, assuming 7nm technology node.
Each CPU instruction incurs an additional ${\sim}13$\,pJ overhead from instruction fetch and register file access, estimated from \cite{horowitz20141} (scaled down to 7nm using SRAM energy ratios from \cite{jouppi2021ten}).
We assume a digital Flash CIM device as in \cite{yu2024cambricon,deng2025kvnand,sun2025lincoln}.
I8 MAC energy is 0.1 pJ \cite{jouppi2021ten}, and
input/output activation buffer energy is 0.8 pJ per 8 bits (extrapolated from 7nm data in \cite{jouppi2021ten}, assuming 2KB I/O buffer as in \cite{yu2024cambricon}).
SSD-CIM SRAM overhead (${\sim}7$\,pJ per 32 bits) is estimated from 1\,MB SRAM read energy \cite{jouppi2021ten}.
FP32 Div $\approx 6\times$ FP32 MAC and FP32 Sqrt $\approx 12\times$ FP32 MAC, estimated from Agner Fog instruction tables for Zen 7\,nm~\cite{fog2025instruction}.
FP32 Exp is typically approximated using ${\sim}8$ FMA instructions~\cite{coremath}, incurring $8\times$ computation and instruction overhead.

\begin{table}[!htbp]
    \caption{
    Per-operation compute energy.
    }\label{tab:hw-compute-energy}
    \centering
    \scriptsize
    \begin{tabular}{llcc}
    \toprule
    \textbf{Device} & \textbf{Operation} & \textbf{Op Energy (pJ)} & \textbf{Overhead (pJ)} \\
    \midrule
    CPU & I32 Add        & 0.03 & ${\sim}13$ \\
    CPU & I32 Mul        & 1.5  & ${\sim}13$ \\
    CPU & FP32 Add       & 0.4  & ${\sim}13$ \\
    CPU & FP32 Mul       & 1.3  & ${\sim}13$ \\
    CPU & FP32 Div       & 10.2 & ${\sim}13$ \\
    CPU & FP32 Sqrt      & 20.4 & ${\sim}13$ \\
    CPU & FP32 Exp       & 13.6 & ${\sim}8 \times 13$ \\
    \midrule
    SSD-CIM & I8 MAC     & 0.01  & ${\sim}0.8$ \\
    SSD-CIM & I32 Mul    & 1.5  & ${\sim}7$ \\
    SSD-CIM & I32 Add    & 0.03 & ${\sim}7$ \\
    SSD-CIM & Argmax     & 0.5  & ${\sim}7$ \\
    \bottomrule
    \end{tabular}
    \end{table}

\paragraph{Memory energy.}
Table~\ref{tab:hw-memory-energy} reports memory access energy per bit.
DRAM read/write energy (7\,pJ/bit) is from \cite{deng2025kvnand}.
SSD energy (7.9\,pJ/bit) and 
internal Flash read energy (3\,pJ/bit) are from \cite{deng2025kvnand}.

\begin{table}[!htbp]
    \caption{
    Memory access energy per bit (read/write).
    }\label{tab:hw-memory-energy}
    \centering
    \scriptsize
    \begin{tabular}{lcc}
    \toprule
    \textbf{Device} & \textbf{Energy/Bit (pJ)} & \textbf{Source} \\
    \midrule
    DRAM  & 7 & \cite{deng2025kvnand} \\
    SSD (Flash) & 7.9 & \cite{deng2025kvnand} \\
    SSD (Flash, internal) & 3.0 & \cite{deng2025kvnand} \\
    \bottomrule
    \end{tabular}
    \end{table}

\paragraph{Bandwidth.}
Table~\ref{tab:hw-bandwidth} lists the memory interface bandwidths assumed in our model: 76.8\,GB/s for DDR5 DRAM (Dual-Channel) and 8\,GB/s for PCIe 4.0.

\begin{table}[!htbp]
    \caption{
    Memory interface bandwidth.
    }\label{tab:hw-bandwidth}
    \centering
    \scriptsize
    \begin{tabular}{lcc}
    \toprule
    \textbf{Interface} & \textbf{Bandwidth} & \textbf{Standard} \\
    \midrule
    CPU $\leftrightarrow$ DRAM & 76.8\,GB/s & DDR5 \\
    CPU $\leftrightarrow$ SSD  & 8\,GB/s    & PCIe 4.0 \\
    \bottomrule
    \end{tabular}
    \end{table}

\paragraph{Capacity.}
Table~\ref{tab:hw-capacity} lists our storage capacity assumptions.
We assume a 512\,GB SSD (32\,GB per die) with 192\,MB of on-chip SRAM (using 6MB/die from \cite{sun2025lincoln}).
We assume 64GB for the DRAM capacity.

\begin{table}[!htbp]
    \caption{
    Storage capacity assumptions.
    }\label{tab:hw-capacity}
    \centering
    \scriptsize
    \begin{tabular}{lcc}
    \toprule
    \textbf{Device} & \textbf{Capacity} & \textbf{Source} \\
    \midrule
    DRAM & 64GB & --- \\
    SSD (Flash)      & 512\,GB (32\,GB per die) & --- \\
    SSD (on-chip SRAM) & 192\,MB                 & \cite{sun2025lincoln} \\
    \bottomrule
    \end{tabular}
    \end{table}

\paragraph{Throughput.}
Table~\ref{tab:hw-throughput} reports peak compute throughput.
CPU throughput (2\,TFLOPS FP32) is based on Zen 2 (7\,nm) and is within the range of listed configurations \cite{suggs2020zen2,amd_epyc7002_datasheet}.
NPU throughput (32\,TFLOPS BF16) assumes a $16\times$ speedup over the CPU with matrix extension, and is consistent with \cite{sun2025lincoln}.
SSD-CIM I8 MAC throughput (1\,TB/s) is from \cite{sun2025lincoln} (determined by peak internal read bandwidth), scaled assuming the capacity from Table~\ref{tab:hw-capacity}.
We assume SSD-CIM peak computation throughput for postprocessing in the peripheral logic is 20x lower than internal CIM throughput (50 GFLOPS).

\begin{table}[!htbp]
    \caption{
    Peak compute throughput.
    }\label{tab:hw-throughput}
    \centering
    \scriptsize
    \begin{tabular}{lcc}
    \toprule
    \textbf{Device} & \textbf{Peak Throughput} & \textbf{Source} \\
    \midrule
    CPU (FP32 MAC)    & 2\,TFLOPS   & \cite{suggs2020zen2,amd_epyc7002_datasheet} \\
    NPU (BF16 MAC)    & 32\,TFLOPS & \cite{sun2025lincoln} \\
    SSD-CIM (I8 MAC)  & 1\,TB/s     & \cite{sun2025lincoln} \\
    SSD-CIM (I32 Mul) & 50\,GFLOPS  & --- \\
    SSD-CIM (I32 Add) & 50\,GFLOPS  & --- \\
    \bottomrule
    \end{tabular}
    \end{table}

\section{Nonlinear Operation Offloading Ablations}
Table \ref{tab:nonlinear-ablation} shows ablations for offloading nonlinear operations to the NPU vs. our approach using integer approximations in the controller across short (1k)/medium (64K)/long (256K) context lengths. Both scenarios assume a streaming implementation of nonlinear operations. Offloading uses the NPU to do the nonlinear operations, and our approach (integer ops) uses the controller to approximate nonlinear operations with integer approximations. 

\begin{table}[!htbp]
    \caption{
    Nonlinear layer forward pass latency (ms) and energy (J) across context lengths for Llama-3.1-8B (batch size 1). Parenthesized values show speedup ($\uparrow$) / energy reduction ($\uparrow$) relative to offloading.
    }\label{tab:nonlinear-ablation}
    \centering
    \scriptsize
    \setlength{\tabcolsep}{3pt}
    \begin{tabular}{ll|cc|cc|cc}
    \toprule
    & & \multicolumn{2}{c|}{\textbf{1K}} & \multicolumn{2}{c|}{\textbf{64K}} & \multicolumn{2}{c}{\textbf{256K}} \\
    & \textbf{Configuration} & Latency & Energy & Latency & Energy & Latency & Energy \\
    \midrule
    \multirow{2}{*}{Softmax} & CIM, offloading  & 0.018 & 6.04e-06 & 0.264 & 9.66e-05 & 4.197 & 1.54e-03 \\
                             & CIM, integer ops & 0.010 (1.78$\times$) & 1.42e-06 (4.25$\times$) & 0.147 (1.80$\times$) & 2.27e-05 (4.25$\times$) & 2.334 (1.80$\times$) & 3.63e-04 (4.25$\times$) \\
    \midrule
    \multirow{2}{*}{RMSNorm} & CIM, offloading  & 2.099 & 7.71e-04 & 33.556 & 1.23e-02 & 536.873 & 1.97e-01 \\
                             & CIM, integer ops & 0.664 (3.16$\times$) & 1.47e-04 (5.25$\times$) & 10.613 (3.16$\times$) & 2.35e-03 (5.25$\times$) & 169.792 (3.16$\times$) & 3.76e-02 (5.25$\times$) \\
    \midrule
    \multirow{2}{*}{SiLU}    & CIM, offloading  & 7.342 & 2.72e-03 & 117.450 & 4.35e-02 & 1879.168 & 6.96e-01 \\
                             & CIM, integer ops & 2.616 (2.81$\times$) & 5.13e-04 (5.30$\times$) & 41.839 (2.81$\times$) & 8.21e-03 (5.30$\times$) & 669.412 (2.81$\times$) & 1.31e-01 (5.30$\times$) \\
    \bottomrule
    \end{tabular}
\end{table}

Table \ref{tab:nonlinear-e2e} shows full system modeling results comparing offloading nonlinear operations to the NPU vs. our approach using integer approximations in the controller across short (1k)/medium (64K)/long (256K) context lengths. Our approach results in speedup and less energy consumption, and scales better as the sequence length increases.

\begin{table}[!htbp]
    \caption{
    System-level decode latency (TBT, ms) and energy (J) across context lengths for Llama-3.1-8B (batch size 1). CIM, offloading refers to offloading nonlinear operations to the NPU; CIM, integer ops is our approach. Parenthesized values show speedup ($\uparrow$) / energy reduction ($\uparrow$) relative to offloading. NL \% shows the fraction of total TBT and energy attributable to nonlinear layers.
    }\label{tab:nonlinear-e2e}
    \centering
    \scriptsize
    \setlength{\tabcolsep}{3pt}
    \begin{tabular}{l|cc|cc|cc}
    \toprule
    & \multicolumn{2}{c|}{\textbf{1K}} & \multicolumn{2}{c|}{\textbf{64K}} & \multicolumn{2}{c}{\textbf{256K}} \\
    \textbf{Configuration} & TBT (ms) & Energy (J) & TBT (ms) & Energy (J) & TBT (ms) & Energy (J) \\
    \midrule
    CIM, offloading  & 54.86 & 37.44 & 111.47 & 39.72 & 283.99 & 46.66 \\
    \quad NL \% of total & 2.1\% & 0.1\% & 30.7\% & 4.0\% & 47.5\% & 13.6\% \\
    \midrule
    CIM, integer ops & 54.53 (1.01$\times$) & 37.41 (1.00$\times$) & 96.48 (1.16$\times$) & 38.49 (1.03$\times$) & 224.33 (1.27$\times$) & 41.80 (1.12$\times$) \\
    \quad NL \% of total & 1.5\% & 0.0\% & 19.9\% & 1.0\% & 33.5\% & 3.6\% \\
    \bottomrule
    \end{tabular}
\end{table}

\section{Full System Results}
\label{sec:appendix-full-system-results}
Table \ref{tab:sys-combined-full} provides full system modeling results across short (1K)/medium (64K)/long (256K) context length configurations.

\begin{table}[!htbp]
    \caption{
    System-level decode latency (TBT, ms) and energy (J) across context lengths for Llama-3.1-8B (batch size 1).
    B1/B2 are NPU+DRAM baselines; S1/S2 are CIM-SSD variants.
    Parenthesized values show speedup ($\uparrow$) / energy reduction ($\uparrow$) relative to B1.
    }\label{tab:sys-combined-full}
    \centering
    \scriptsize
    \setlength{\tabcolsep}{3pt}
    \begin{tabular}{ll|cc|cc|cc}
    \toprule
    & & \multicolumn{2}{c|}{\textbf{1K}} & \multicolumn{2}{c|}{\textbf{64K}} & \multicolumn{2}{c}{\textbf{256K}} \\
    & \textbf{Configuration} & TBT & Energy & TBT & Energy & TBT & Energy \\
    \midrule
    B1 & NPU, no compr.             & 183.7 & 101.8 & 294.3 & 164.2 & 631.5 & 354.5 \\
    B2 & NPU + dict.\ compr.        & 190.4 (0.96$\times$) & 108.2 (0.94$\times$) & 197.1 (1.49$\times$) & 114.9 (1.43$\times$) & 217.4 (2.91$\times$) & 135.1 (2.62$\times$) \\
    \midrule
    S1 & CIM-SSD, coeff.\ on ctrl.  & 54.1 (3.40$\times$)  & 37.4 (2.72$\times$)  & 92.2 (3.19$\times$)  & 38.3 (4.29$\times$)  & 208.4 (3.03$\times$) & 40.9 (8.66$\times$)  \\
    S2 & CIM-SSD, coeff.\ on DRAM   & 59.9 (3.07$\times$)  & 37.8 (2.69$\times$)  & 80.6 (3.65$\times$)  & 41.4 (3.96$\times$)  & 143.5 (4.40$\times$) & 52.5 (6.75$\times$)  \\
    \bottomrule
    \end{tabular}
    \end{table}